\documentclass[letterpaper, 10 pt, conference]{ieeeconf}  

\IEEEoverridecommandlockouts                              

\usepackage{graphicx}
\usepackage{amsmath} 
\usepackage{amssymb}  
\usepackage{multirow}
\newcommand{\D}[2]{D_{#2}{#1}}

\newcommand{\cantor}{\mathrm{cantor}}
\newcommand{\skewsym}[1]{{\left\lfloor #1 \right\rfloor_\times}} 

\title{\LARGE \bf
LiPO: LiDAR Inertial Odometry for ICP Comparison
}

\author{Darwin Mick$^{1}$, Taylor Pool$^{2}$, Madankumar Sathenahally Nagaraju$^{2}$, Michael Kaess$^{2}$, \\ Howie Choset$^{2}$, and Matt Travers$^{2}$  
\thanks{$^{1}$Darwin Mick is from the Department of Mechanical Engineering, Carnegie Mellon University, USA., $^{2}$Taylor Pool, Madankumar Sathenahally Nagaraju, Michael Kaess, Howie Choset, and Matt Travers are from the Robotics Institute, Carnegie Mellon University, USA.
        {\tt\small \{dmick, tpool, msathena, kaess, choset, mtravers\}@andrew.cmu.edu}}%
}

\begin{document}
\maketitle
\thispagestyle{empty}
\pagestyle{empty}

\begin{abstract}

We introduce a LiDAR inertial odometry (LIO) framework, called LiPO, that enables direct comparisons of different iterative closest point (ICP) point cloud registration methods.
The two common ICP methods we compare are point-to-point (P2P) and point-to-feature (P2F).
In our experience, within the context of LIO, P2F-ICP results in less drift and improved mapping accuracy when robots move aggressively through challenging environments when compared to P2P-ICP.
However, P2F-ICP methods require more hand-tuned hyper-parameters that make P2F-ICP less general across all environments and motions.
In real-world field robotics applications where robots are used across different environments, more general P2P-ICP methods may be preferred despite increased drift.
In this paper, we seek to better quantify the trade-off between P2P-ICP and P2F-ICP to help inform when each method should be used.
To explore this trade-off, we use LiPO to directly compare ICP methods and test on relevant benchmark datasets as well as on our custom unpiloted ground vehicle (UGV).
We find that overall, P2F-ICP has reduced drift and improved mapping accuracy, but, P2P-ICP is more consistent across all environments and motions with minimal drift increase.

\end{abstract}

\section{Introduction}\label{sec.intro}
Point-to-point iterative closest point (P2P-ICP) is a standard technique used in robot localization and mapping.
P2P-ICP point cloud registration finds correspondences among points, as opposed to features, while being more general to different sensors than point-to-feature (P2F) methods.  
In fact, P2P-ICP for LiDAR odometry has seen a resurgence in the state estimation community with KISS-ICP \cite{vizzo2023kiss} demonstrating comparable results against hand-tuned P2F methods while using less hyper-parameters. 
However, when the robot experiences high accelerations, constant velocity assumptions inherent to both P2P and P2F-ICP are violated and both methods will experience severe drift.
To address high accelerations, prior work developed LiDAR-inertial odometry (LIO) to use inertial measurement units (IMUs) to augment ICP performance \cite{zhang2014loam}.

In our experience, LIO methods that use P2F-ICP (P2F-LIO) see decreased drift and improved mapping accuracy in difficult environments during aggressive motion when compared to P2P-LIO.
However, P2F-LIO uses many hyper-parameters for feature extraction, and as such, is more difficult to generalize across environments than P2P-LIO.
Hence, despite the potential for increased drift, P2P-LIO may still be preferred in field robotics applications where robots are used across many different environments. 
It is both practical and useful to understand the magnitude of the drift difference between P2P-LIO and P2F-LIO to better understand when each should be used. 
In this paper, we demonstrate performance differences in terms of drift and mapping accuracy between P2P-LIO and P2F-LIO across different environments and motions to highlight the advantages and disadvantages of each method.

\begin{figure}
    \centering
    \includegraphics[width=0.9\linewidth]{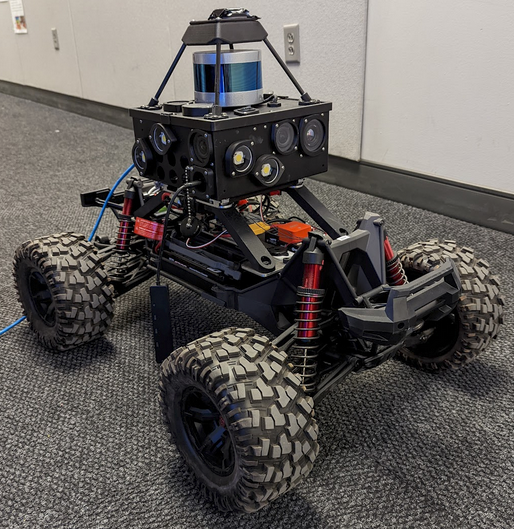}
    \caption{Unpiloted ground vehicle (UGV) equipped with time synchronized VLP-16 LiDAR, EPSON G366 IMU, cameras, and NVIDIA ORIN. Note, only LiDAR and IMU sensors are used for state estimation.}
    \label{fig:RC_Car}
\end{figure}

In order to highlight differences between P2P-LIO and P2F-LIO, we introduce a modular LIO system, LiPO. With LiPO we can directly compare ICP methodologies while maintaining consistency across all other aspects of the algorithm. 
To do so, we take inspiration from Super Odometry \cite{zhao2021super} in our IMU-Centric LIO approach.
Our P2P-ICP module also takes inspiration from KISS-ICP and our P2F module follows the P2F-ICP method described in Super Odometr y~\cite{zhao2021super,vizzo2023kiss}. 
Our IMU bias estimation follows the factor graph framework common in many LIO methods~\cite{shan2020lio}. 
We demonstrate the efficacy of our method on the M2DGR~\cite{yin2021m2dgr} and UrbanNav~\cite{hsu2021urbannav} datasets and on an unpiloted ground vehicle (UGV) (Fig.~\ref{fig:RC_Car}) moving aggressively in different environments (Fig.~\ref{fig:paths}). 
Finally, using our UGV, we demonstrate lower drift and improved mapping accuracy using P2F-LIO while P2P-LIO is more consistent across all environments and motions with minor increases in drift.

We structure the remainder of this paper as follows: Section~\ref{section:related_work} discusses prior work and explains its relevance to the work presented. Section~\ref{section:Background} gives a brief mathematical background required for our work.
Section~\ref{section:methodology} details the approach and how we built a LIO system.
Section~\ref{section:experiments} explains how we benchmarked LiPO and details the robotic platform we used for testing.
Section~\ref{section:results} explicitly compares estimation accuracy and mapping results between our P2P-LIO and P2F-LIO frameworks. 
Section~\ref{section:conclusion} concludes the paper. 

\section{Related Work}\label{section:related_work}
Finding the rigid body transformation that best aligns two point clouds is the core enabling technology behind LiDAR odometry.
Besl~\cite{besl1992method} proposed a method, commonly called Iterative Closest Point (ICP), that achieved this objective by repeatedly matching points between the two clouds, then minimizing the transform between those matches until convergence.
Chen~\cite{chen1992object} proposed a similar algorithm, but replaced the point-to-point metric with a point-to-plane variant to speed up registration by modeling the surface of interest.
Zhang~\cite{zhang1994iterative} proposed detecting bad associations using their respective distances, which increased robustness.
Generalized ICP \cite{segal2009generalized} expanded Chen's point-to-plane metric with a probabilistic weighting based on the spread of points forming the surface of interest.
LiDAR Odometry and Mapping (LOAM) \cite{zhang2014loam} exploited the structure of the LiDAR scanner pattern to detect geometrically reliable planes and edges.
While other registration methods exist, such as Fast Global Registration \cite{koltenfastglobal}, they fall outside of the scope of this paper because they do not rely on ICP.
LiDAR odometry methods such as LOAM \cite{zhang2014loam}, F-LOAM \cite{wang2021f}, LIO-SAM \cite{shan2020lio}, LEGO-LOAM \cite{shan2018lego}, and Super Odometry \cite{zhao2021super} utilize feature-based registration.

Recent LiDAR odometry methods have focused on robustness.
A main method to do so is via robust cost functions, which stems from the computer vision commmunity \cite{zhang1997parameter}.
KISS-ICP~\cite{vizzo2023kiss} adapts the kernel of a robust cost function according to the correction magnitude of previous ICP iterations.

Due to the sequential nature of LiDAR point returns, a scan collected at high speeds relative to the surrounding environment will be stretched along the direction of motion.
This stretching, or skewing, of the scan presents a real challenge.
In the absence of any other sensor, modeling the LiDAR scanner as moving at a constant velocity over the course of the scan allows for deskewing the point cloud \cite{vizzo2023kiss}.
Unfortunately, the presence of significant accelerations during the scan degrades the velocity estimate.
Methods such as LOAM \cite{zhang2014loam} introduced an inertial measurement unit to counteract this effect.

LiDAR-Inertial Odometry address this problem by fusing registration data with the IMU itself.
Conventionally, Kalman filtering methods \cite{kalmanfilter} \cite{xu2021fast, xu2022fast, he2023point, zhang2024msc} have been the \emph{defacto} choice for fusing registration and IMU data.
Unfortunately, these approaches commit to linearization points during marginalization, thus resulting in a sub-optimal strategy for nonlinear problems.
More recently, methods such as LIO-SAM \cite{shan2020lio} have utilized a sliding window approach to optimize over more states to reduce linearization errors.

\section{Background}\label{section:Background}

\subsection{3D Rigid Body Transformations}
The space of 3D rigid body transformations forms a Lie group, which is a group that is also a smooth manifold.
Sol\`a \cite{sola2018micro} contains an excellent introduction into Lie theory.
Here, we cover some basics and explain notation used throughout the rest of the paper.
Let $T$ denote a rigid body transformation.
Then $T$ is composed of two items: 1) a rotation, $\mathcal{R}$, 2) a translation, $t$.
$\mathcal{R}$ is also a Lie Group.
We use $\circ$ to denote the group operator, and $*$ to denote the action $\mathcal{R}$ and $T$ perform on elements of $\mathbb{R}^3$, whether by rotation or translation.
Many representations exist for rotations including orthogonal matrices of determinant 1, quaternions, and axis-angle.
Similarly, many representations exist for rigid body transformations including homogeneous matrices, quaternion vector pairs, and dual quaternions.

\subsection{Differentiation}
We use Spivak \cite{spivak2018calculus} notation for derivatives.
Given Lie Groups $\mathcal{X}, \mathcal{Y}$ with tangent spaces of dimensions $n$ and $m$ respectively, a function $f: \mathcal{X} \rightarrow \mathcal{Y}$, and operators $\oplus: \mathcal{X} \times \mathbb{R}^n \rightarrow \mathcal{X}$ and $\boxminus: \mathcal{Y} \times \mathcal{Y} \rightarrow \mathbb{R}^m$, the Fr\'echet derivative of the function evaluated at $x \in \mathcal{X}$ is defined as $Df(x) \in \mathbb{R}^{m \times n}$ such that
\begin{IEEEeqnarray}{lCl}
    \lim_{\xi \rightarrow 0} \frac{|| f(x \oplus \xi) \boxminus f(x) - Df(x) ||}{|| \xi ||} & = & 0
\end{IEEEeqnarray}
We use $\D{f}{x}(x,y)$ to denote the partial derivative of $f$ with respect to $x$.

\subsection{IMU Model}
We model the IMU according to \cite{forster2016manifold}.
Let $\omega$ denote angular velocity in the IMU frame. We model the measured angular velocity $\hat{\omega}$ as corrupted by a slowly evolving bias, $b_\omega$, and a zero-mean white noise, $\eta_\omega$.
Formally,
\begin{IEEEeqnarray}{lCl}
    \hat{\omega} & = & \omega + b_\omega + \eta_\omega \\
    \dot{b}_\omega & \sim & \mathcal{N}\left( 0, \Sigma_{b_\omega} \right) \\
    \eta_\omega & \sim & \mathcal{N}\left(0, \Sigma_{\eta_\omega} \right)
\end{IEEEeqnarray} 

Additionally, let $a$ denote linear acceleration in the IMU frame.
Let $g$ denote gravity in the world frame.
The IMU measures specific force, $f$, which is the acceleration added with the normal force due to gravity.
We also model the specific force as corrupted by a bias, $b_a$, and zero-mean white noise, $\eta_a$.
\begin{IEEEeqnarray}{lCl}
    f & = & a - \mathcal{R}^{-1} \circ g \\
    \hat{f} & = & f + b_a + \eta_a \\
    \dot{b}_a & \sim & \mathcal{N}\left( 0, \Sigma_{b_a} \right) \\
    \eta_a & \sim & \mathcal{N}\left( 0, \Sigma_{\eta_a} \right)
\end{IEEEeqnarray}

Overall, the bias term $b$ represents the ordered pair of biases $(b_\omega, b_a)$.

\section{Methodology}\label{section:methodology}

\subsection{Overview} 
\begin{figure}
    \centering
    \includegraphics[width=0.85\linewidth]{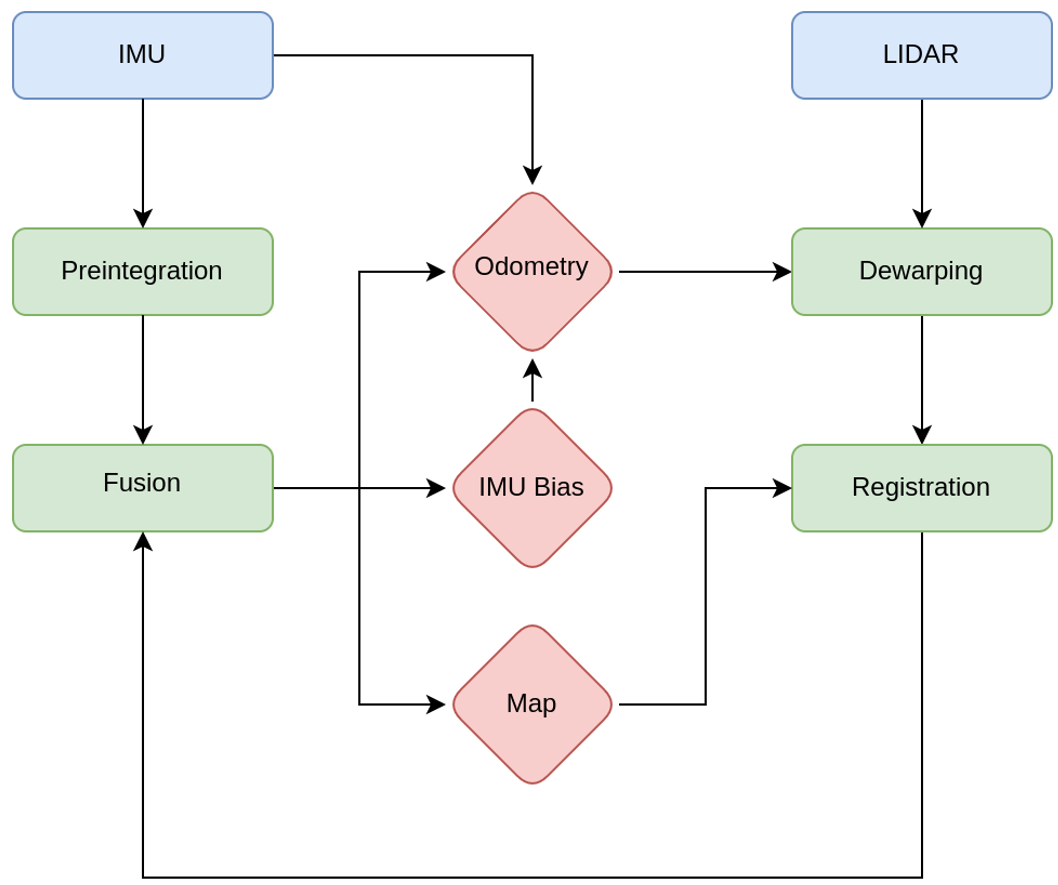}
    \caption{Full system architecture; Blue cells are inputs; Green cells are intermediate components; Red cells are outputs}
    \label{fig:enter-label}
\end{figure}

We describe the full system in conjunction with Figure~\ref{fig:enter-label}. 
First, when we receive a scan, we append it to a buffer.
We also receive and forward integrate IMU measurements for a high rate odometry.
When all odometry measurements during the time of the scan are received, we proceed to transform each point of the scan into the world frame using the odometry.
Next, we use this newly transformed point cloud to perform registration, either with features, or via our point-to-point method.
Once the registration is complete, we have obtained a hypothesis as to the pose of the IMU in the world frame at the time of the scan.
Using this hypothesis, we fuse with preintegrated IMU measurements in a factor graph.
After optimizing the factor graph over the poses, velocities, and IMU biases associated with each frame, we then add the points from the scan into the map using the smoothed pose information.

\subsection{Dewarping}

Given a simple radial LiDAR scanner, each return occurs at a slightly different time.
If the sensor is in motion during data collection, this time delta will cause a naive accumulation of points without accounting for timing to be distorted.
A common approach to compensate for this distortion is to transform each point according to the time it was received into one common frame.
Methods to compute the transform for each point are varied--the simplest approach being the application of the exponential map for a given estimate of robot velocity over the course of the scan \cite{sola2018micro}.
Unfortunately, we observed that this approach, while appealing in its simplicity, did not accurately capture the motion of the sensor, and thus did not succeed to undistort at high magnitudes of acceleration.
To compensate for this, we employed higher rate pose information (produced by IMU motion propagation).
We found that a first order interpolation scheme to find the pose of the LiDAR sensor at the time of the point return was sufficient.
We note that this approach requires strict time synchronization of the IMU and LiDAR sensor, as well as requiring all poses derived from the IMU relevant to the points in a LiDAR scan to be stored in a buffer before processing the scan itself.

\subsection{Voxel Grid}
Our work leverages a voxel grid to perform efficient nearest-neighbor search and downsampling.
Another key benefit is that parallelization is easy to perform across voxel cells.
Specifically, we choose a voxel width, then we create a hash function.
There are many choices for hash functions.
We chose a function that uses the Cantor hash, which is reversible.
We extend it to three dimensions.
\begin{IEEEeqnarray}{lCl}
x, y, z \in \mathbb{N} & &  \\
\cantor\left( x, y \right) & = & \frac{(x + y)(x+y+1)}{2} + y \\
\mathrm{hash}\left( x, y, z \right) & = & \mathrm{cantor}\left(x, \mathrm{cantor}\left(y, z \right)\right)
\end{IEEEeqnarray}

We hash each voxel cell by coordinates determined by dividing each point location by the voxel dimension, and then performing a floor operation, then cast to an unsigned integer $\in \mathbb{N}$.

\subsection{Point-to-Point Implementation}
The point to point registration algorithm proceeds as follows: we have a
set of points, $X$ to be registered against another set of points, $Y$.
We seek for a rigid body transformation, expressing frame $X$ in frame $Y$, ${^Y T_X }$.
We use an iterative method, and we assume first that
$X$ has been transformed according to the high rate update or the constant velocity model presented earlier.
Then, for each iteration, for each $x_i \in X$, we find the closest $y_i \in Y$.
The $ith$ residual function we seek to minimize is the following:

\begin{IEEEeqnarray}{lCl}
   r(T) & = & T*x - y 
\end{IEEEeqnarray}

In this case,
\begin{IEEEeqnarray}{lCl}
Dr & = & \D{\left[T * x \right]}{T}
\end{IEEEeqnarray}

\subsubsection{Geman-McClure}

Unfortunately, the probability of bad associations is significant.
If left unchecked, these incorrect matches can give rise to large residuals, which will adversely affect the registration.
To combat this behavior, we down-weight larger squared residuals by a Geman-McClure robust kernel \cite{zhang1997parameter}.
The Geman-McClure kernel is given by
\begin{IEEEeqnarray}{lCl}
\rho(y) & = & \frac{y}{y + \kappa} 
\end{IEEEeqnarray}

Using this residual, we have
\begin{IEEEeqnarray}{lCl} 
f_i \left( x \right) & = & \frac{1}{2} \rho\left( r_i^\mathrm{T} (x) r_i (x) \right)
\end{IEEEeqnarray}

With this residual, we can leverage the chain rule to obtain the following derivative of the overall cost function:
\begin{IEEEeqnarray}{lCl}
D\rho D^\mathrm{T} r Dr + D^2 \rho D^\mathrm{T} y Dy &&
\end{IEEEeqnarray}

\subsection{Point-to-Plane Implementation}

A planar feature is given by three points $y_1, y_2, y_3$, and a corresponding point $x$ to be matched. We form the residual function as follows:

Let $n$ be the unit normal vector created from $y_1, y_2, y_3$.
Then $n = \mathrm{normalize} \left( (y_2 - y_1) \times (y_3 - y_1) \right)$.
Let $u$ be the vector pointing from the plane to $x$.
$u(T_x, T_y) = (T_x * x - T_y * y_1)$.
\begin{IEEEeqnarray}{lCl}
	r(T_x, T_y) & = & (\mathcal{R}_y * n)^\mathrm{T} u
\end{IEEEeqnarray}

We compute the derivative of the residual function as
\begin{IEEEeqnarray}{lCl}
	\D{r}{} & = & (\mathcal{R}_y * n)^\mathrm{T} \D{u}{} + u^\mathrm{T} \D{[\mathcal{R}_y * n]}{T_y}
\end{IEEEeqnarray}

Now, note that
\begin{IEEEeqnarray}{lCl}
	\D{u}{} & = & \begin{bmatrix}
		\D{[T_x * x]}{T_x} & -\D{[T_y * y_1]}{T_y}
	\end{bmatrix}
\end{IEEEeqnarray}

Also,
\begin{IEEEeqnarray}{lCl}
	\D{[R_y * n]}{T_y} & = & \begin{bmatrix}
		\D{[R_y * n]}{R_y} & 0_{3 \times 3}
	\end{bmatrix}
\end{IEEEeqnarray}

\subsection{Point-to-Edge Implementation}

Let $y_1, y_2 \in \mathbb{R}^3$ form an edge line to be matched with a point $x \in \mathbb{R}^3$. 
Let $u(T_y) = R_y * \mathrm{normalized}\left( y_2 - y_1 \right)$.
Let $x^\prime(T_x) = T_x * x - T_y * y_1$.
Then the edge residual function is
\begin{IEEEeqnarray}{lCl}
    r(T_x, T_y) & = & x^\prime \times u
\end{IEEEeqnarray}

The derivative of the residual function is
\begin{IEEEeqnarray}{lCl}
    Dr & = & \begin{bmatrix}
        -\skewsym{u} \D{x^\prime}{} & \skewsym{x^\prime} \D{u}{}
    \end{bmatrix}
\end{IEEEeqnarray}

\subsection{IMU Bias Factor}
\label{sec:imu_bias_est}

We model the IMU bias $b = (b_\omega, b_a)$ as Brownian motion \cite{evans2012introduction}.

We assume that there is no correlation between $b_\omega$ and $b_a$, so that we define the covariance of $b_{k+1}$ given $b_k$ as 
\begin{IEEEeqnarray}{lCl}
    \Sigma_{b_{k+1/k}} & = & \begin{bmatrix}
        \Sigma_{b_\omega} & 0 \\
        0 & \Sigma_{b_a}
    \end{bmatrix} \Delta t_{k+1/k}
\end{IEEEeqnarray}
Note that the covariance scales according to the time difference between $t_{k+1}$ and $t_k$.

The residual is simply the difference between the two biases:
\begin{IEEEeqnarray}{lCl}
    r(b_{k+1}, b_k) & = & b_{k+1} - b_k
\end{IEEEeqnarray}

\subsubsection{Preintegrated IMU Factor}

Between each LiDAR scan, we accumulate IMU measurements into a buffer.
We integrate IMU measurements from the pose $T_k$ and velocity $v_k$ of the previous LiDAR scan to time of the next LiDAR scan according to the following equations:
\begin{IEEEeqnarray}{lCl}
    \mathcal{R}_{k+1} & = & \mathcal{R}_k \circ \mathrm{Exp}\left( \omega_k \Delta t_{k+1/k} \right) \\
    p_{k+1} & = & p_k + v_k \Delta t_{k+1/k} \\
    & & + \frac{1}{2} (g + \mathcal{R}_k * f_k) \Delta t_{k+1/k}^2 \\
    v_{k+1} & = & v_k + (g + \mathcal{R}_k * f_k) \Delta t_{k+1/k}
\end{IEEEeqnarray}

A naive implementation of the IMU factor would have to reintegrate the measurements for each change in the initial conditions $T_k, v_k, b_k$.
However, we eliminate this inefficiency via the technique described in \cite{forster2016manifold} for on-manifold IMU preintegration.

\subsubsection{ICP Factor (Absolute Pose)}

We provide an absolute pose constraint from the registration, because we perform the registration in the global frame.

\begin{IEEEeqnarray}{lCl}
    r(T) & = & \mathrm{Log}\left( T_\mathrm{absolute}^{-1} \circ T \right)
\end{IEEEeqnarray}

\subsubsection{Smoothing}
\begin{figure}
    \centering
    \includegraphics[width=0.5\linewidth]{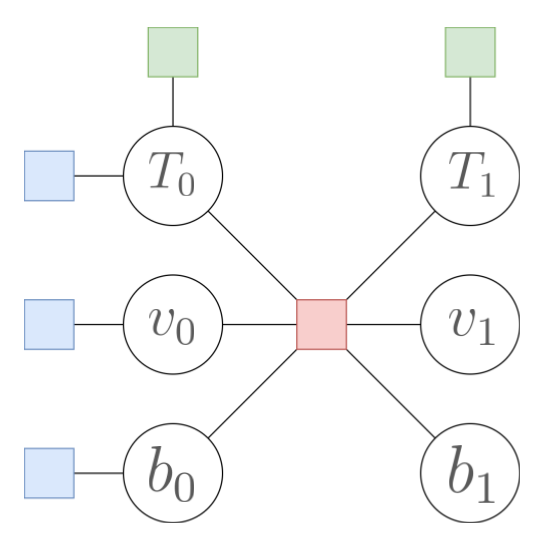}
    \caption{First two frames of factor graph showing prior factors (blue), absolute pose factors (green), combined preintegration and IMU bias factor (red)}
    \label{fig:smoothing}
\end{figure}

In smoothing, we combine a preintegration factor with an absolute pose factor to create a factor graph.
Figure \ref{fig:smoothing} displays the smoothing formulation in terms of the factor graph.
A factor graph is a bipartite graph consisting of two types of nodes called variables and factors.
The factors represent cost functions over the variables.
Solving the factor graph shown is equivalent to minimizing a cost function, which in our case is the maximum a posteriori.
We provide values for the start of the optimization routine from prior beliefs and odometry propagation.

\begin{IEEEeqnarray}{lCl}
    \arg \min_{T,v,b} & & \mathrm{prior}(T_0) + \mathrm{prior}(v_0) + \mathrm{prior}(b_0) \\
    & & + \sum_{i=0}^n \mathrm{icp}\left( T_i \right) \\ &&
    + \sum_{i=0}^{n-1} \mathrm{imu} \left( T_i, v_i, b_i, T_{i+1}, v_{i+1}, b_{i+1} \right)
\end{IEEEeqnarray}

\subsection{System Parameters}

Table~\ref{tab:preintegration} lists IMU preintegration, while Table~\ref{tab:absolute_pose_factor} lists parameters for our global pose factor.
Note, we found empirically that different hyper-parameters between P2P and P2F were required to maximize the performance of LiPO.
We tuned these hyper-parameters once, then left them consistent for all of the results presented in the paper.

\begin{table}
\centering
\caption{Preintegration Parameters}
\begin{tabular}{|| c | c ||}
\hline
\bf{Name} & \bf{Value} \\
\hline
    Accelerometer variance & 1.07e-2 \\
    Gyroscope variance & 1.92e-3 \\
    Accelerometer bias variance & 5.27e-9 \\
    Gyroscope bias variance & 8.12e-9 \\
\hline
\end{tabular}
\label{tab:preintegration}
\end{table}

\begin{table}
    \centering
    \caption{Absolute Pose Factor Parameters}
    \begin{tabular}{|| c | c ||}
    \hline
    \bf{Name} & \bf{Value} \\
    \hline
        Translation standard deviation (P2F) & 1.89e-2 \\
        Rotation standard deviation (P2F) & 1.21e-2 \\
        Translation standard deviation (P2P) & 1.90e-1 \\
        Rotation standard deviation (P2P) & 1.22e-1 \\
    \hline
    \end{tabular}
    \label{tab:absolute_pose_factor}
\end{table}

\section{Experiments}
\label{section:experiments}
In this section we detail the benchmark datasets that we use to validate our algorithm's performance.
Additionally, we describe our custom data set.
We compare our methods to a common and open sourced P2P-LIO method, FAST-LIO2\cite{xu2022fast}, and a common and open sourced P2F-LIO method, LIO-SAM\cite{shan2020lio}.
We refer to our method as P2P-LiPO and P2F-LiPO for our method that uses P2P-ICP and P2F-ICP respectively.

\subsection{Multi-sensor and Multi-scenario SLAM Dataset for Ground Robots (M2DGR)}
We use the Multi-sensor and Multi-scenario SLAM Dataset for Ground Robots (M2DGR) to validate our approach \cite{yin2021m2dgr}.
This system uses a Velodyne-32C LiDAR as well as Handsfree A9 IMU.
The ground truth is collected with GNSS-RTK.
We compare our P2P and P2F-based LIO algorithms against FAST-LIO2 and LIO-SAM \cite{xu2022fast,shan2020lio}.
We leverage the street01-street10 data and report our performance on the average of these data and use the hyper-parameters reported in M2DGR for these algorithms. 

\subsection{UrbanNav DataSet}
Additionally, we use the UrbanNav Data Set to test our approach \cite{hsu2021urbannav}.
This system records data using a car driving through busy city streets.
Data is recorded with a HDL 32E Velodyne as well as as an Xsense Mti 10 IMU.
Although the vehicle is additionally equipped with a slanted LiDARs, we do not use them.
Ground truth is collected with a NovAtel SPAN-CPT GNSS-IMU receiver.
We focus on UrbanNav HK-Whampoa and UrbanNav HK-TST.
We keep hyper-parameters consistent between this dataset and M2DGR.

\subsection{Custom Dataset}
We leverage our custom robotic UGV platform as shown in Figure~\ref{fig:RC_Car}.
The UGV is equipped with a VLP-16 LiDAR, an EPSON-G366 IMU, and an NVIDIA Orin.
We collect ground truth data using a Leica Total Station.
Further details of the full UGV system can be found in Sringanesh et al. \cite{sriganesh2024modular}.
Although we post-process data to collect results in our paper, we note that P2P-LiPO and P2F-LiPO are capable of real-time performance on our system.

\begin{figure}
    \centering
    \includegraphics[width=1\linewidth]{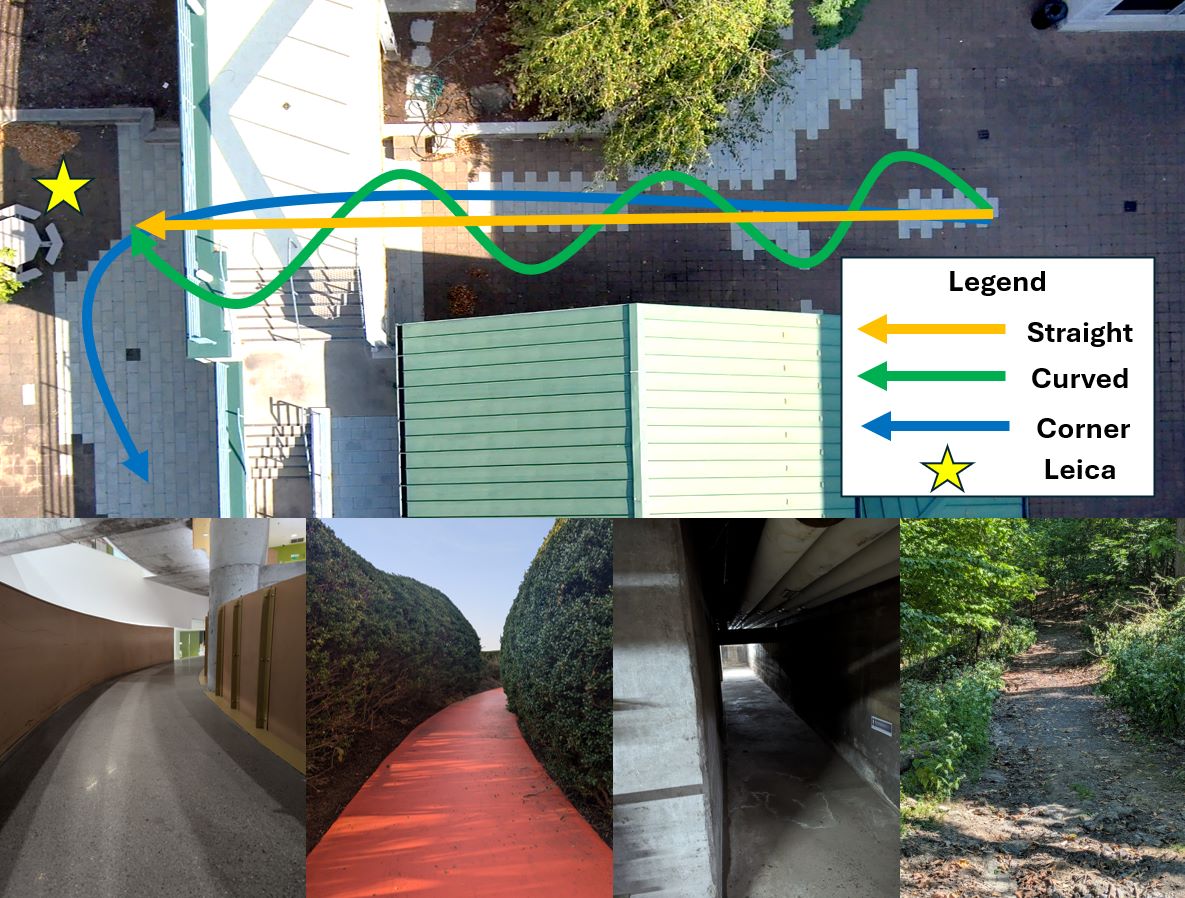}
    \caption{Testing environments from our custom data set. Where the top image demarcates the paths driven where we collected ground truth from the Leica station. Then, we show examples of our challenging environments. From left to right, we have loop, number garden, catacombs, park.}
    \label{fig:paths}
\end{figure}

We leverage three different paths, with increasing speed.
Path 1, is a straight line, path 2 is a curved line, and path 3 is a path where the robot turns a corner shown in Figure~\ref{fig:paths}.
For the straight and curved line paths we drive the car at 2, 4, and 6 m/s and record 5 runs at each speed.
For the corner path, we run the car at 2 and 4 m/s due to physical limitations of our system, and take 5 runs at each speed.

Additionally, we test in four environments that we have found to be particularly challenging for our robot.
These environments are ``catacombs", ``number garden", ``loop", and ``park".
All three of these environments require the robot to maneuver tight and constrained spaces that consistently have geometrically ambiguous features that are known to harm estimation performance.
Plus, due to the small environment, the robot is prone to collisions, and trials where the robot collides with the environment are not removed.
Example pictures of the environments can be seen in Figure~\ref{fig:paths}. We ran the robot through each of these environments ten times at 2 m/s and we started and stopped the robot from the same position to measure final position drift, since there is not persistent line-of-sight to use the Leica Total Station.

\section{Results} \label{section:results}
First, we compare the average relative pose error (RPE) and average absolute pose error (APE) between the ground truth path and estimated path for all 4 methods across each benchmark dataset and paths from our custom dataset with ground truth. 
For the sake of space, we average the ATE and RPE for all tests in each benchmark datasets, M2DGR and UrbanNav. We use the evo\_traj library and its Umeyama alignment functionality to calculate these results \cite{grupp2017evo}. 
Results are tabulated in Table~\ref{table:Results1}.

\begin{table}
\centering
\caption{Average Absolute Pose Error (APE) and Average Relative Pose Error (RPE) for all Environments and Methods}
\label{table:Results1}
    \begin{tabular}{|| c | c c c||} 
     \hline
     \textbf{Sequence} & \textbf{Method} & \textbf{Avg. ATE} & \textbf{Avg. RPE} \\ [0.5ex] 
     \hline\
     \multirow{4}{4em}{M2DGR} 
        & FAST-LIO2 & 1.40 & 0.256  \\ 
        & LIO-SAM & 7.49 & 0.037  \\ 
        & P2P-LiPO & 1.69 & \textbf{0.018} \\
        & P2F-LiPO & \textbf{1.44} & 0.021 \\
     \hline
     \multirow{4}{4em}{UrbanNav} 
        & FAST-LIO2 & 5.97 & 0.064  \\ 
        & LIO-SAM & 33.7 & 2.41  \\ 
        & P2P-LiPO & \textbf{4.13} & \textbf{0.103} \\
        & P2F-LiPO & 15.3 & 0.196 \\
     \hline
     \multirow{4}{4em}{Straight} 
        & FAST-LIO2 & \textbf{0.135} & 0.970  \\ 
        & LIO-SAM & 0.989 & 0.722  \\ 
        & P2P-LiPO & 0.952  & \textbf{0.643} \\
        & P2F-LiPO & 0.930 & 0.693 \\
     \hline
     \multirow{4}{4em}{Curved} 
        & FAST-LIO2 & \textbf{0.560} &  0.708 \\ 
        & LIO-SAM & 0.919 & 0.737  \\ 
        & P2P-LiPO & 0.861 & \textbf{0.679} \\
        & P2F-LiPO & 0.800 & 0.702 \\
     \hline
     \multirow{4}{4em}{Corner} 
        & FAST-LIO2 & 1.414 & 0.832  \\ 
        & LIO-SAM & 0.705 & 0.498  \\ 
        & P2P-LiPO & \textbf{0.660} & 0.481 \\
        & P2F-LiPO & \textbf{0.660} & \textbf{0.480} \\  [1ex] 
     \hline
    \end{tabular}
\end{table}

\begin{figure}
    \centering
    \includegraphics[width=\linewidth]{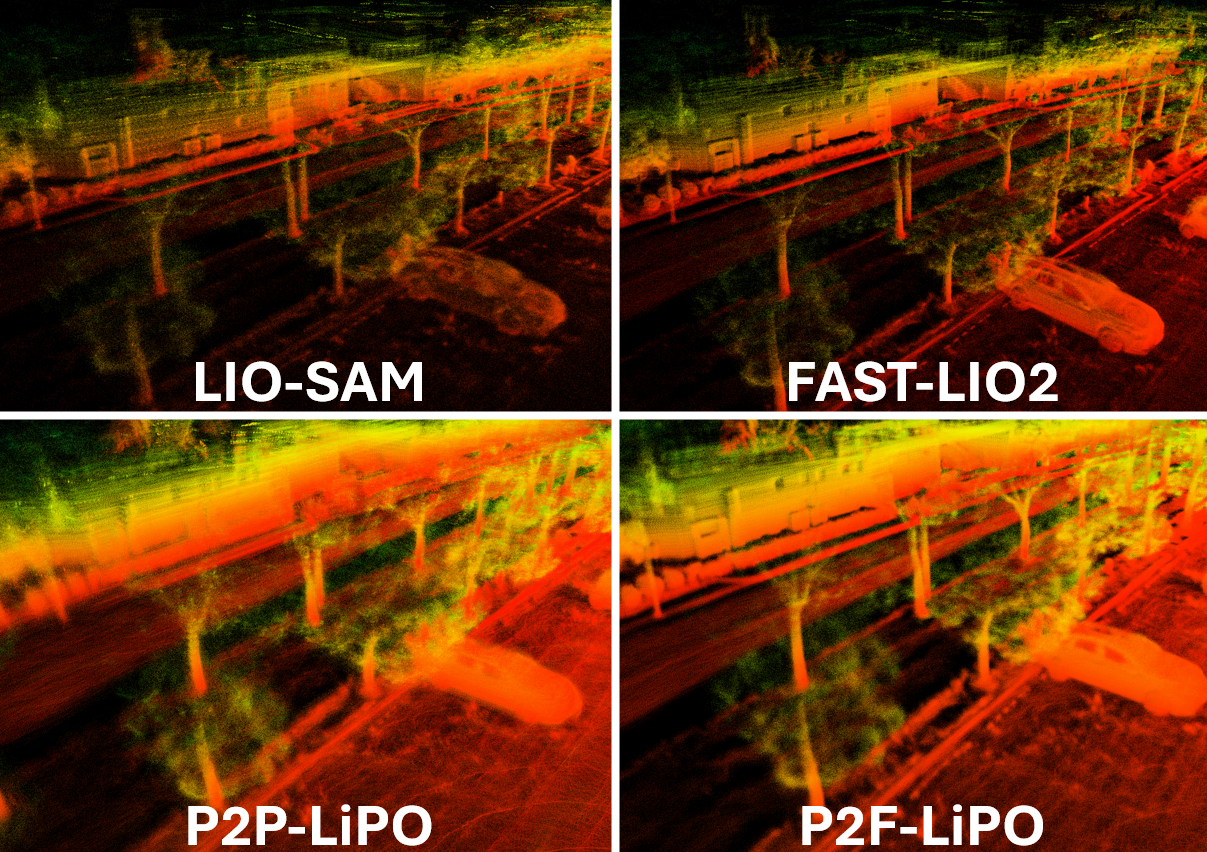}
    \caption{Map comparison between LIO-SAM, FAST-LIO2, P2P-LiPO, and P2F-LiPO. We observe the similarity in all maps, where the car, the curbs, and the trees are well defined. We do note that the background of P2P-LiPO is noticeably less defined than the other methods.}
    \label{fig:map_comparison}
\end{figure}

We can see both LiPO methods have competitive performance against Fast-LIO2 and LIO-SAM in all environments. Additionally, we show in Figure~\ref{fig:map_comparison} the qualitative mapping performance of LiPO compared to LIO-SAM and FAST-LIO2 from street\_3 from the M2DGR dataset. These results validate the performance of LiPO both emprically and qualitatively. Next, we calculate average final drift error between P2P-LiPO and P2F-LiPO on our four challenging datasets and report the results in Table~\ref{table:env_results}.

\begin{table}
\centering
\caption{Average Final Position Drift for Challenging Environments}
\label{table:env_results}
    \begin{tabular}{|| c | c c ||} 
     \hline
     \textbf{Sequence} & \textbf{Method} & \textbf{Avg. Final Position Drift} \\ [0.5ex] 
     \hline\
     \multirow{2}{8em}{Catacombs} 
        & P2P-LiPO & 0.410  $\pm$ 0.083\\
        & P2F-LiPO & \textbf{0.202 $\pm$ 0.117}\\
     \hline
     \multirow{2}{8em}{Number Garden} 
        & P2P-LiPO & 2.21 $\pm$ 1.59  \\
        & P2F-LiPO & \textbf{1.86 $\pm$ 1.39} \\
     \hline
     \multirow{2}{8em}{Loop} 
        & P2P-LiPO & 0.522 $\pm$ 0.377 \\
        & P2F-LiPO & \textbf{0.209 $\pm$ 0.090} \\ 
     \hline
      \multirow{2}{8em}{Park} 
        & P2P-LiPO &  0.200 $\pm$ 0.106  \\
        & P2F-LiPO & \textbf{0.100 $\pm$ 0.051 } \\ [1ex] 
     \hline
    \end{tabular}
\end{table}

\begin{figure}
    \centering
    \includegraphics[width=\linewidth]{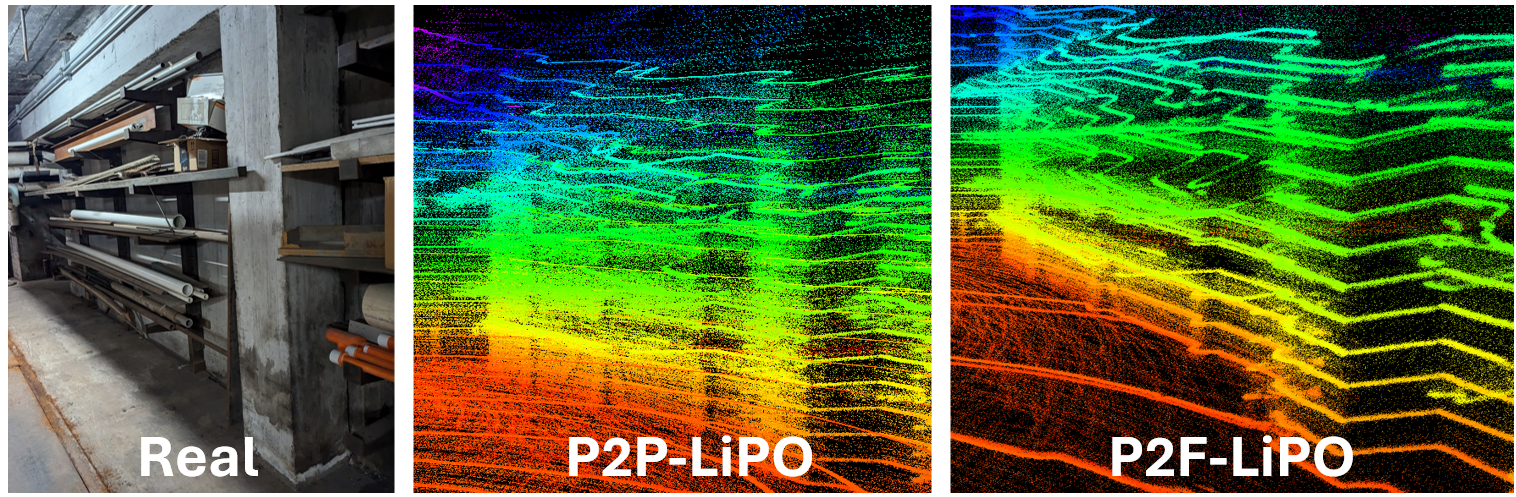}
    \caption{Map comparison between P2P-LiPO and P2F-LiPO, where we can qualitatively observe map smearing in the P2P-LiPO case but not for P2F-LiPO. The shelves, pillar, and floor are much more defined in the P2F-LiPO map.}
    \label{fig:catacomb}
\end{figure}

P2F-LiPO outperforms P2P-LiPO in every case in the challenging environments, where P2F-LiPO has performance increases by up to 30 cm. Plus, mapping results are clearer in the P2F-LiPO version as detailed in Figure~\ref{fig:catacomb}. Across all datasets, P2F-LiPO tends to outperform P2P-LiPO with the exception of the UrbanNav dataset, where both feature-based methods, P2P-LiPO and LIO-SAM, significantly under perform. Since all testing was done with the same hyper parameter tuning, it is likely that the feature-based methods failed due to a poor tune. In no data set, did the P2P-LiPO perform as poorly as P2F-LiPO did on UrbanNav. As such, depending on the system use case, the consistent performance of P2P-LIO methods may be preferred over the increased accuracy but less general performance of P2F-LIO.

\section{Conclusion}
\label{section:conclusion}

In this paper, we have introduced and detailed our own LIO framework, LiPO, and validated its performance against FAST-LIO2 and LIO-SAM. Additionally, we have demonstrated that P2F-LIO outperforms P2P-LIO in challenging environments with aggressive motion. In future work, we will delve deeper into the causes of these differences and seek to further improve accuracy of P2P-LiPO to work towards a method that is both highly accurate and generalizable.

\addtolength{\textheight}{-12cm}   




\section*{Acknowledgements}

We thank Shibo Zhao for productive and meaningful conversations regarding this work. We thank Sahil Chaudhary for his work in developing and maintaining our hardware platform.

\bibliographystyle{IEEEtran}
\bibliography{IEEEabrv,citations}
\end{document}